\documentclass[runningheads]{llncs}

\usepackage{eccv}

\usepackage{eccvabbrv}

\usepackage{graphicx}
\usepackage{booktabs}
\usepackage[table]{xcolor}

\usepackage[accsupp]{axessibility}  

\definecolor{cvprblue}{rgb}{0.21,0.49,0.74}
\PassOptionsToPackage{colorlinks=true,linkcolor=cvprblue,citecolor=cvprblue,urlcolor=blue}{hyperref}
\usepackage{hyperref}

\usepackage{orcidlink}

\definecolor{myyellow}{HTML}{ffd83d}

\definecolor{citecolor}{HTML}{0071bc}

\def\method{LoHo-Manip\xspace}

\newcommand{\myparagraph}[1]{\vspace{-0.1mm}\noindent\textit{#1}}

\usepackage{titletoc}
\providecommand{\authcount}[1]{}

\newcommand\blfootnote[1]{%
  \begingroup
  \renewcommand\thefootnote{}\footnote{#1}%
  \addtocounter{footnote}{-1}%
  \endgroup
}

\begin{document}

\title{Long-Horizon Manipulation via Trace-Conditioned VLA Planning} 

\titlerunning{Lo-Ho Manip}

\author{
Isabella Liu\inst{1}
\and
An-Chieh Cheng\inst{1}
\and
Rui Yan\inst{1}
\and
Geng Chen\inst{1}
\and
Ri-Zhao Qiu\inst{1}
\and
Xueyan Zou\inst{1}
\and
Sha Yi\inst{1}
\and
\text{ }Hongxu Yin\inst{2}\textsuperscript{\textdagger}
\and
\text{ }Xiaolong Wang\inst{1}\textsuperscript{\textdagger}
\and
\text{ }Sifei Liu\inst{2}\textsuperscript{\textdagger}}

\authorrunning{I.~Liu et al.}

\institute{University of California, San Diego \and NVIDIA}
\maketitle

\vspace{-2em}
\begin{figure}[h]
\centering
\includegraphics[width=1\textwidth]{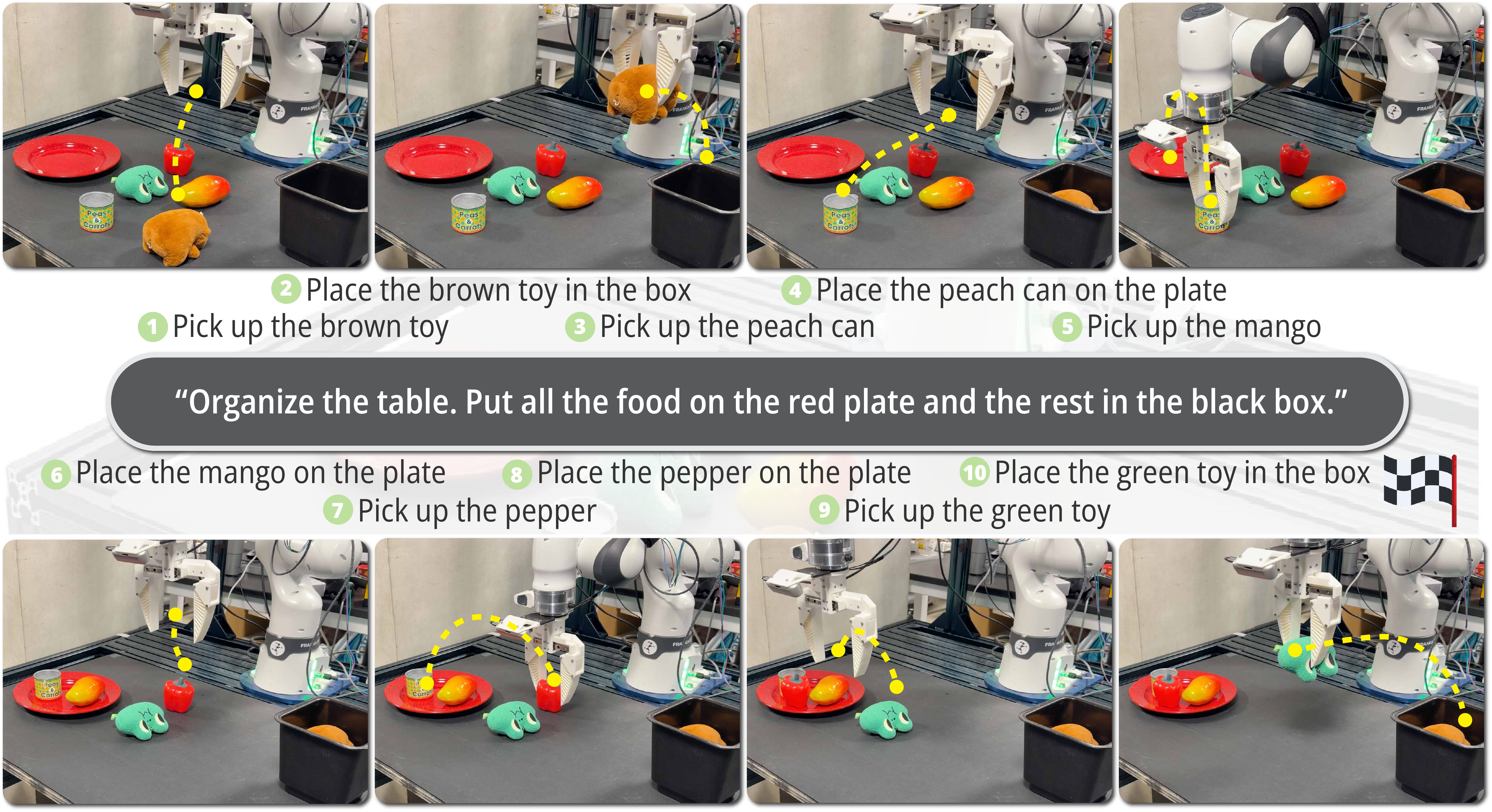}
\vspace{-1.5em}
\caption{\footnotesize Our task manager decouples high-level planning from low-level control by decomposing a high-level instruction into sequential sub-tasks and tracking their completion over time. For each step, it predicts a spatial visual trace that acts as an actionable prompt for a low-level VLA executor, enabling reliable execution of complex multi-step manipulation tasks. Project page: \href{https://www.liuisabella.com/LoHoManip/}{\textcolor{cvprblue}{https://www.liuisabella.com/LoHoManip}}.}
\vspace{-2em}
\label{fig:teaser}
\end{figure}

\blfootnote{\footnotesize \textsuperscript{\textdagger} Equal advising.}

\vspace{-1em}
\abstract{
Long-horizon manipulation remains challenging for vision-language-action (VLA) policies: real tasks are multi-step, progress-dependent, and brittle to compounding execution errors. We present \method, a modular framework that scales short-horizon VLA execution to long-horizon instruction following via a dedicated task-management VLM. The manager is decoupled from the executor and is invoked in a receding-horizon manner: given the \emph{current} observation, it predicts a progress-aware \emph{remaining plan} that combines (i) a subtask sequence with an explicit \emph{done + remaining} split as lightweight language memory, and (ii) a visual trace---a compact 2D keypoint trajectory prompt specifying where to go and what to approach next. The executor VLA is adapted to condition on the rendered trace, thereby turning long-horizon decision-making into repeated local control by ``following the trace.'' Crucially, predicting the remaining plan at each step yields an implicit closed loop: failed steps persist in subsequent outputs, and traces update accordingly, enabling automatic continuation and replanning without hand-crafted recovery logic or brittle visual-history buffers. Extensive experiments spanning embodied planning, long-horizon reasoning, trajectory prediction, and end-to-end manipulation in simulation and on a real Franka robot demonstrate strong gains in long-horizon success, robustness, and out-of-distribution generalization. 
\vspace{-0.5em}
\vspace{-0.5em}
}
\section{Introduction}
\label{sec:intro}
\vspace{-1em}

Robots have made striking progress in short-horizon manipulation---picking, placing, opening, and pushing---driven by large-scale imitation learning~\cite{rt1,rt2,openx,chi2025diffusion} and increasingly capable vision-language(-action) models~\cite{pi0,lbm,gr00t-n1,openvla}. Yet the gap between these skills and long-horizon manipulation remains wide~\cite{longvla,liu2024bidirectional}. Real tasks rarely end after a single grasp. They require dozens of interdependent to-dos, state-dependent decisions (e.g., ``the kettle is not full enough''), and robust recovery when any step fails. A household instruction such as ``refill the kettle'' implicitly expands into a long to-do list: locate a cup, grasp it, move to the faucet, fill it to an appropriate level, close the faucet, navigate to the kettle, open the lid, pour, close the lid, and so on. Long-horizon success demands not only low-level control but also \emph{task management}: decomposition, progress tracking, and continuous re-evaluation as the world changes.

A common strategy is to increase the capability of a single monolithic policy—either by scaling a VLA model~\cite{rt1,rt2,pi0}, or by embedding high-level planning within the same model that produces actions~\cite{palme,pi0-5,thinkact,cotvla}. However, tightly coupling planning and execution introduces two persistent limitations. First, fragility under drift~\cite{dagger,florence2022implicit,janner2022planning}: long sequences amplify small errors, and one-shot plans cannot reliably anticipate partial failures, occlusions, or object motion. Second, poor modularity~\cite{devin2017learning,kumar2021rma,cheng2024navila}: if the planner is fused into the executor, upgrading or swapping the low-level VLA (e.g., changing embodiments, action spaces, or training domains) typically requires re-engineering the entire stack. This tension is especially acute in embodied settings, where task complexity, environment variability, and distribution shift are the norm.

We argue that long-horizon manipulation benefits from a cleaner separation of concerns: a high-level component should focus on what remains to be done, while a low-level component should focus on how to do the next short-term control. Concretely, we introduce \method, a hierarchical system centered around a dedicated task-management VLM that sits above any VLA policy. The manager is responsible for task-level reasoning and outputs two complementary artifacts at inference time:
\begin{itemize}
    \item \textit{Remaining subtasks}: a structured list describing the actions that still need to be completed from the current observation onward.
    \item \textit{Visual trace}: a high-level 2D trajectory (a ``roadmap'') that specifies where the agent should move and which objects/regions should be approached next.
\end{itemize}
The trace is not merely an auxiliary visualization; it is a conditioning signal for the executor. By training or adapting the VLA to follow this trace, we convert a difficult long-horizon planning problem into a sequence of short-horizon control problems, where modern VLAs excel. Intuitively, the manager answers ``\textit{what/where next}'' with an explicit visual pointer, while the VLA handles ``\textit{how to do it}'' with precise actions.

A key design choice in \method is the use of \emph{receding-horizon} task management via \emph{remaining-plan prediction}. Instead of generating a complete plan only at the first frame, the manager is invoked periodically (or at every step) and always predicts the remaining plan from the current observation. This yields a simple but powerful closed-loop property: if a subtask fails (e.g., the cup was not grasped), the world state reflects that failure, and the manager continues to include the uncompleted item in its subsequent plan — often accompanied by an updated trace. As a result, \method exhibits implicit progress tracking, implicit replanning, and implicit recovery without requiring hand-crafted failure detectors or special-case logic.

Importantly, long-horizon management must represent \emph{history} without becoming brittle or expensive. Feeding long observation histories can increase latency and expose the manager to distribution shift when execution becomes imperfect, while emitting only a single ``next subtask'' can lead to unstable instruction streams under repeated failures. \method adopts a lightweight alternative: the manager conditions solely on the \emph{current} observation, while maintaining progress through a compact \emph{textual memory} that summarizes what has already been completed (e.g., ``done: a--c; remaining: d--f''). This simple design simultaneously preserves explicit task-state bookkeeping, avoids reliance on long visual histories, and keeps the subtask interface stable by predicting a \emph{remaining} sequence rather than a one-step directive.

As the additional visual prompt to the VLA executor, the trace also offers a direct path to generalization. Many VLAs overfit to the visual and semantic statistics of their training distributions (e.g., a particular set of objects, textures, or colors). In contrast, a general-purpose VLM is often better at grounding language to unseen objects and producing a coarse spatial plan. The trace externalizes this grounding as a visual instruction (e.g., literally drawing a path toward an object of an unseen category). Once the executor learns the generic skill of trace-following, it can execute behaviors on targets it has never encountered, provided the manager can point to them.

We evaluate \method on multiple long-horizon embodied and manipulation benchmarks and show that our framework effectively bridges high-level reasoning with low-level control. Our experiments, spanning embodied planning, long-horizon reasoning, trajectory prediction, and end-to-end manipulation in both simulation and on a real Franka robot, demonstrate strong gains in long-horizon success, robustness, and out-of-distribution generalization. By decomposing complex instructions into grounded sub-task trajectories, our model maintains temporal consistency and significantly outperforms existing vision-language baselines in both simulated and real-world tabletop environments. We summarize the contributions as follows:
\begin{itemize}
    \item We propose \method, a modular framework that separates a dedicated task-management VLM from a short-horizon VLA executor, enabling reusable high-level management across different low-level policies.
    \item We train the manager to predict \emph{remaining} subtasks (and the corresponding trace) from the current observation throughout an episode, yielding implicit progress tracking, re-planning, and failure recovery without hand-crafted heuristics.
    \item We introduce a visual trace as an actionable prompt for the executor, turning long-horizon planning into local control and improving robustness and generalization to novel objects and environments.
\end{itemize}

\begin{figure}[t]
\centering
\includegraphics[width=1\textwidth]{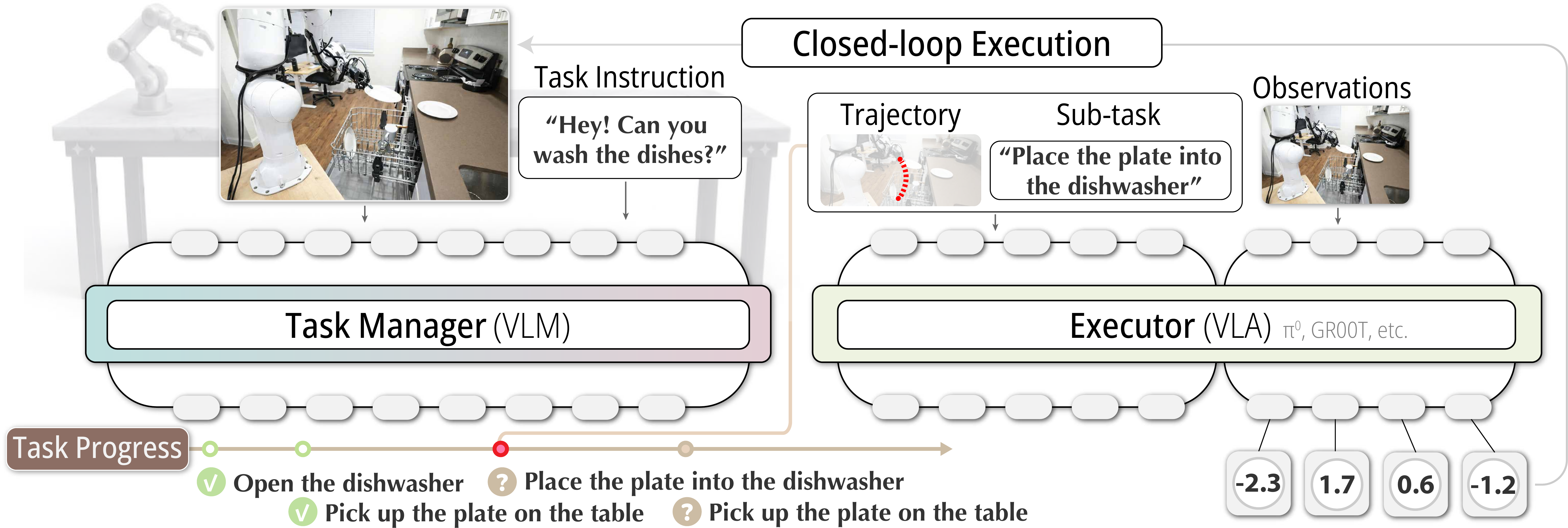}
\vspace{-1em}
\caption{\footnotesize \textbf{Overview of \method framework.}
Given a task instruction and the current observation, a vision–language \emph{task manager} predicts the next sub-task together with a spatial visual trace indicating the intended interaction trajectory. A low-level \emph{executor} then performs short-horizon control conditioned on this guidance. The system operates in a \emph{closed loop}: after executing actions, new observations are fed back to the manager to update task progress and generate the next sub-task and trace, enabling robust long-horizon manipulation and recovery from execution errors.}
\vspace{-2em}
\label{fig:method}
\end{figure}

\vspace{-1em}
\section{Method}
\vspace{-0.5em}
\label{sec:method}

Directly learning a policy that maps observations to long-horizon action sequences
is difficult in practice: the policy must implicitly track task progress,
handle compounding execution errors, and maintain consistent behavior over many steps.
As the horizon grows, these challenges lead to unstable credit assignment and
distribution shift between training trajectories and real executions.

To address this, we adopt a hierarchical decomposition that separates
\emph{task management} from \emph{short-horizon control}.
The manager predicts the remaining high-level plan from the current observation,
while a low-level policy executes locally conditioned actions.
This formulation converts long-horizon manipulation into sequential short-horizon
execution with explicit progress signals.

\vspace{-1em}
\subsection{System Overview}
\label{sec:system}

Our system decomposes long-horizon manipulation into two components:
a high-level \textbf{task manager} that reasons about task progress,
and a low-level \textbf{executor} that performs short-horizon control.

\myparagraph{Task Manager.}
The task manager is a vision-language model that predicts the
\emph{remaining task structure} from the current observation.
Given the instruction $x$ and the current observation $o_t$,
the manager produces two outputs:
(1) a \textbf{subtask description} $s_t$ that specifies the next
short-horizon objective in natural language, and
(2) a \textbf{visual trace} $\tau_t$ represented as a sequence of
2D keypoints indicating the spatial trajectory or interaction
target associated with the subtask.
Importantly, the task manager is designed as a \emph{generalist module}.
It does not depend on the specific low-level policy used for execution,
allowing the same manager to interface with different VLA models.

\myparagraph{Executor (VLA).}
The executor is a standard action policy
(e.g., $\pi_0$, GR00T, or other VLA/VA backbones) that maps observations
to robot actions.
Modern VLA models are particularly effective at executing
\emph{short-horizon instructions}, making them well suited to
follow the subtasks produced by the manager.

To connect the two components, the predicted visual trace
$\tau_t$ is provided as an additional input to the executor.
The trace is rendered into the observation and used as a spatial
conditioning signal.
After fine-tuning on this representation, the VLA learns to
follow the predicted trajectory and execute the corresponding
subtask.

\myparagraph{Closed-loop execution.}
During execution, the system repeatedly invokes the task manager
to predict the remaining plans from the current observation.
The executor then performs locally conditioned actions guided by the predicted subtask and trace. This receding-horizon interaction enables explicit task progress
tracking and robust recovery from execution errors.

\begin{figure}[t]
\centering
\includegraphics[width=1\textwidth]{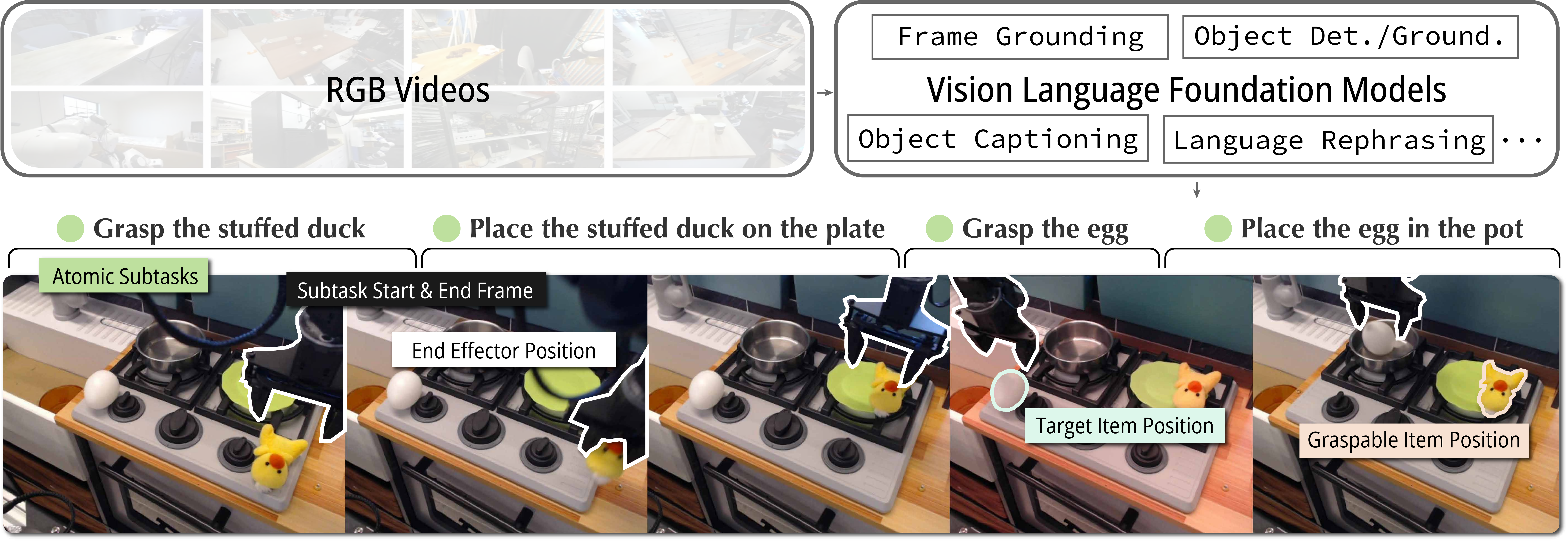}
\vspace{-1.5em}
\caption{\textbf{Data Pipeline.} Given RGB manipulation videos, we use vision–language foundation models to perform frame grounding, object detection, and captioning to identify interaction events. The trajectory is segmented into atomic subtasks with corresponding start and end frames, while the robot end-effector positions are extracted to form a 2D visual trace. Additional grounding identifies target and graspable objects, producing paired (subtask instruction, visual trace) supervision for training the task manager.}
\label{fig:data_pipeline}\vspace{-1em}
\end{figure}

\begin{figure}[t]
\centering
\includegraphics[width=1\textwidth]{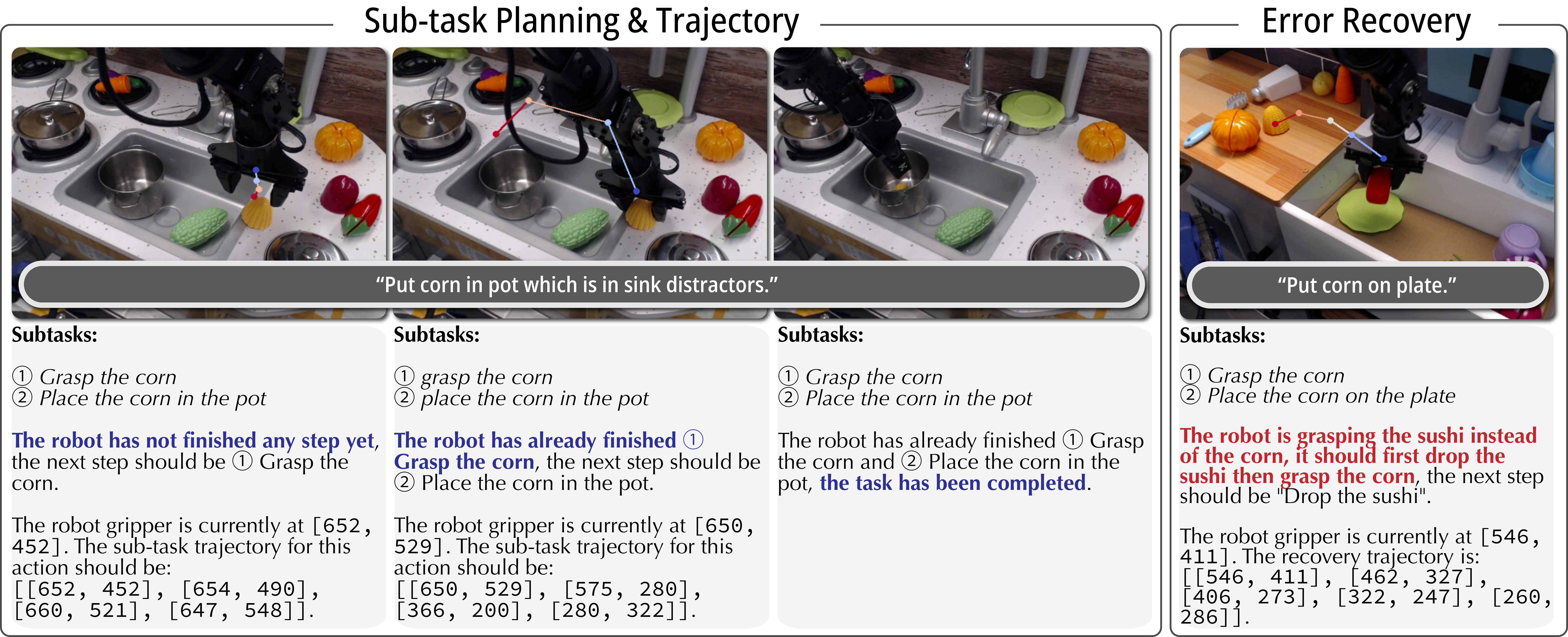}
\vspace{-2em}
\caption{\footnotesize\textbf{Samples of Sub-task Planning and Error Recovery.} 
\textit{Left:} Sequential samples from a single episode demonstrating sub-task decomposition and trajectory grounding. The planning module maintains temporal context to track progress, while predicted trajectories map instructions to spatial coordinates to reduce ambiguity. 
\textit{Right:} A curated error recovery sample. The model identifies a semantic error (grasping sushi instead of corn), triggers a corrective ``Drop the sushi'' sub-task, and generates a recovery trajectory to reset the state and resume the original goal.}
\label{fig:pipeline_samples}\vspace{-1em}
\end{figure}

\vspace{-1em}
\subsection{Sub-Task and Trace Construction} \label{sec:method_trace}
\myparagraph{Progress-aware plan representation.}
Given an instruction $x$ and a trajectory of observations $\{o_t\}_{t=1}^T$,
we represent the task as a sequence of \emph{atomic interaction primitives}
$\bar{S}=[\bar{s}^{(1)},\ldots,\bar{s}^{(K)}]$.
At each time $t$, instead of emitting only a single ``next subtask'',
we define a \emph{progress-aware plan text} that explicitly includes both
the completed prefix and the remaining suffix:
\begin{equation}
    C_t^\star = [\bar{s}^{(1)}, \ldots, \bar{s}^{(k(t)-1)}], \quad
    R_t^\star = [\bar{s}^{(k(t))}, \ldots, \bar{s}^{(K)}],
\end{equation}
where $k(t)$ denotes the index of the current (or next) primitive at time $t$.
We use $C_t^\star$ as a compact \emph{language memory} summarizing what has been done,
and $R_t^\star$ as the remaining plan to be executed.

\myparagraph{Visual trace representation.}
We additionally associate the plan with a lightweight \emph{visual trace} that encodes spatial intent.
For each timestep, we obtain the 2D pixel coordinate of the robot end-effector
$p_t \in \mathbb{R}^2$.
We define the remaining trace label as the future portion of the end-effector trajectory:
\begin{equation}
    \tau_t^\star = \{p_t, p_{t+1}, \ldots, p_{t_K^e}\},
\end{equation}
where $t_K^e$ is the end index of the final primitive.
In practice, $\tau_t^\star$ is stored in a compact form (e.g., resampled waypoints) and rendered as a visual prompt.

\myparagraph{Current-frame conditioning.}
Although supervision is constructed from full trajectories, our task manager is conditioned on the
\emph{current frame only} during both training and inference. Concretely, the manager receives
$(x, o_t, C_{t-1})$ and predicts $(C_t, R_t, \tau_t)$.
This keeps the visual input distribution consistent between training and deployment, while $C_t$
provides explicit progress context without requiring history frames.

\myparagraph{Constructing primitives and traces.}
To construct $\bar{S}$ and $\{\tau^{(k)}\}$ from video trajectories, we (i) temporally segment the episode into
interaction events and map each segment to an atomic primitive, and (ii) extract end-effector pixel locations
to form traces.
For segmentation, we leverage off-the-shelf vision--language models with video grounding capabilities to identify
physically grounded interaction events and their temporal spans $(t_k^{s}, t_k^{e})$.
For trace extraction, we localize the robot end-effector per frame to obtain $p_t$,
which directly yields $\tau^{(k)}$ for each segment.
In simulation, the end-effector location can be obtained from state; on real robots, it can be obtained
via visual localization.
We defer model-specific choices and implementation details to Sec.~\ref{sec:training}.

\myparagraph{Shift-resilient language memory.}
A practical challenge in long-horizon management is representing history without introducing train--test mismatch.
Feeding long visual histories can be brittle when real executions deviate from smooth demonstrations,
leading to distribution shift in the history frames (also noted in~\cite{mem}).
We therefore condition the manager on the \emph{current} frame only, and encode history as a compact textual summary ($C_t$: ``completed primitives'') alongside the remaining plan ($R_t$).
This preserves explicit progress context for decision making, avoids reliance on history frames, and keeps inference efficient.

\vspace{-1em}
\subsection{Training Paradigm}
\label{sec:training}

Our training separates the \emph{task manager} from the \emph{executor}.
The manager is trained to predict a progress-aware \emph{remaining plan} and its associated trace
from the \emph{current} observation, while the executor is adapted to follow the trace prompt for short-horizon control.
This decoupling keeps the manager executor-agnostic and allows reusing the same manager checkpoint across different VLA backbones,
while each executor is fine-tuned to the trace interface.

\myparagraph{Training data.}
We use two sources of supervision.
(i) \textbf{Real-robot demonstrations} from the Bridge subset (in Open X-Embodiment format) provide video trajectories
from which we extract atomic interaction primitives and end-effector traces following Sec.~\ref{sec:method_trace}~\cite{openx}.
(ii) \textbf{Auxiliary long-horizon reasoning/planning data} (RoboVQA and EgoPlan-BenchIT) provides additional instruction understanding
and progress-reasoning signals that improve the manager's generalization beyond the manipulation distribution~\cite{sermanet2024robovqa,mcarthur2022egoplan}. 
To improve the task manager’s robustness to execution errors, we augment the training set with synthesized failure-recovery samples from the Bridge dataset. We filter for grasp-and-place episodes and identify the transition frames where grasping and placing occur. We then compose "fake" failure data by replacing the original grasped object with other detected graspable items in the scene. A fail recovery example is demonstrated in Fig.~\ref{fig:pipeline_samples}.

\myparagraph{Task manager training.}
We initialize the task manager from a pretrained VLM~\cite{qwen3vl}.
We freeze the vision encoder and fine-tune the language model with supervised learning to predict:
(a) a progress-aware plan text that includes both completed and remaining primitives, and
(b) the associated 2D trace representation (e.g., waypoint sequence) for the remaining execution.
Crucially, the manager is conditioned on the \emph{current frame only} during training, with history provided via the textual progress summary,
which avoids relying on long visual histories that may drift under imperfect rollouts.

\myparagraph{Executor adaptation.}
For the low-level executor, we adopt the $\pi_{0.5}$ architecture~\cite{pi0-5} and initialize from its base checkpoint.
We then fine-tune the executor to condition on the rendered trace prompt (and optionally the current subtask text),
so that it can reliably track the manager-provided trajectory and execute short-horizon interaction primitives.

\myparagraph{Inference loop and memory update.}
At inference time, we run a receding-horizon closed loop.
At each step (or every fixed interval), the manager takes the current observation and the textual progress summary and predicts an updated
(completed, remaining) plan plus trace.
The executor then acts for a short horizon conditioned on the trace prompt.
We update the textual memory by taking the latest completed/remaining split from the manager output, without feeding history frames.
This design preserves explicit progress context while remaining efficient and robust to distribution shift in long visual histories.

\vspace{-1em}
\section{Experiments}
\vspace{-1em}
\label{sec:exp}

Our evaluation validates \method across four key dimensions. We first verify the model's fundamental reasoning and trajectory prediction capabilities (Sec.~\ref{sec:exp:general}). We then assess the model's sub-task planning accuracy on embodied agent benchmarks (Sec.~\ref{sec:exp:agent}). Moving to closed-loop control, we evaluate end-to-end manipulation in simulation (Sec.~\ref{sec:exp:vla-sim}) to analyze the integration of high-level reasoning and low-level control. We conclude with real-world experiments (Sec.~\ref{sec:exp:vla-real}), showcasing the model's ability to handle long-horizon instructions in unconstrained environments.

\begin{table*}[t]
\small
\centering
\setlength{\tabcolsep}{0.1pt} 
\scalebox{0.97}{ 
{\fontsize{8pt}{9pt}\selectfont
\begin{tabular}{l *{10}{>{\centering\arraybackslash}p{27pt}}} 
\toprule
 & \multicolumn{5}{c}{\textbf{RoboVQA} $\uparrow$} & \multicolumn{5}{c}{\textbf{EgoPlan2} $\uparrow$} \\
\cmidrule(lr){2-6} \cmidrule(lr){7-11}
Methods & B-1 & B-2 & B-3 & B-4 & Avg. & Day. & Wrk. & Rec. & Hob. & Avg. \\
\midrule
\rowcolor{gray!10}
\multicolumn{11}{l}{ \textit{Proprietary Models (API)}}\\
GPT-4V~\cite{achiam2023gpt} & 32.2 & 26.5 & 24.7 & 23.9 & 26.8 & 36.7 & 27.7 & 33.9 & 32.5 & 32.6 \\
Gemini-2.5-Flash~\cite{comanici2025gemini} & 39.1 & 31.6 & 22.9 & 22.1 & 28.9 & 44.2 & 42.3 & 43.2 & 39.1 & 42.4 \\
Gemini-3.0-Flash~\cite{gemini3flash} & 46.6 & 36.4 & 33.8 & 32.2 & 37.3 & 55.8 & 45.7 & 45.9 & 46.3 & 48.8 \\
\midrule
\rowcolor{gray!10}
\multicolumn{11}{l}{ \textit{Open-source Models}}\\
InternVL2.5-2B~\cite{chen2024expanding} & 36.6 & 33.7 & 31.0 & 29.4 & 32.7 & 30.9 & 27.8 & 28.6 & 33.1 & 30.1 \\
InternVL3-2B~\cite{zhu2025internvl3} & 34.4 & 33.9 & 33.5 & 33.3 & 33.8 & 36.9 & 29.9 & 35.6 & 31.5 & 33.4 \\
NVILA-2B~\cite{liu2025nvila} & 38.7 & 34.3 & 31.1 & 29.2 & 33.3 & 34.6 & 26.7 & 33.3 & 31.6 & 31.4 \\
Qwen2.5-VL-3B~\cite{qwen2-5vl} & 42.5 & 36.3 & 28.7 & 31.8 & 34.8 & 29.0 & 27.0 & 30.2 & 28.9 & 28.5 \\
Qwen3-VL-4B~\cite{qwen3vl} & 62.5 & 52.2 & 47.0 & 43.2 & 51.2 & 32.3 & 31.2 & 33.8 & 28.9 & 31.3 \\
Qwen3-VL-8B~\cite{qwen3vl} & 72.6 & 62.9 & 56.4 & 51.4 & 60.8 & 39.6 & 37.7 & 38.8 & 31.6 & 36.6 \\
\midrule
\rowcolor{gray!10}
\multicolumn{11}{l}{ \textit{Embodied Foundation Models}}\\
Magma-8B~\cite{yang2025magma} & 38.6 & 31.5 & 28.1 & 26.7 & 31.2 & 32.1 & 25.7 & 34.4 & 29.3 & 29.8 \\
RoboBrain2.0-3B~\cite{robobrain2-0} & 54.4 & 47.7 & 43.1 & 41.0 & 46.5 & 45.3 & 37.6 & 45.9 & 39.7 & 41.8 \\
RoboBrain2.0-7B~\cite{robobrain2-0} & 37.4 & 31.0 & 27.1 & 25.8 & 30.0 & 39.4 & 29.7 & 33.9 & 32.2 & 33.2 \\
RoboBrain2.5-8B~\cite{robobrain2-5} & 48.8 & 38.2 & 33.2 & 29.8 & 37.5 & 43.5 & 40.8 & 38.8 & 40.1 & 41.2 \\
Fast-ThinkAct-3B~\cite{fastthinkact} & 70.1 & 63.0 & 57.2 & 53.0 & 60.8 & 50.3 & 44.3 & 46.4 & 43.2 & 46.4 \\
ThinkAct-3B~\cite{thinkact} & 62.4 & 57.3 & 52.0 & 49.6 & 55.3 & 46.6 & 41.4 & 45.9 & 42.5 & 44.0 \\
ThinkAct-7B~\cite{thinkact} & 69.1 & 61.8 & 56.0 & 52.4 & 59.8 & 50.1 & 49.8 & 44.8 & 45.2 & 48.2 \\
RynnBrain-8B~\cite{rynnbrain} & 74.3 & 63.9 & 57.3 & 52.6 & 62.1 & 38.9 & 32.6 & 35.5 & 31.4 & 34.8 \\
\rowcolor{myyellow!60}
\textbf{\method-4B} & \textbf{75.1} & \textbf{65.0} & \textbf{58.6} & \textbf{53.5} & \textbf{63.1} & \textbf{60.4} & \textbf{56.4} & \textbf{51.9} & \textbf{54.8} & \textbf{56.7} \\
\bottomrule
\end{tabular}}}
\vspace{0.5em}
\caption{\footnotesize Performance comparison across two key benchmarks: RoboVQA~\cite{sermanet2024robovqa} for long-horizon reasoning and EgoPlan-Bench2~\cite{qiu2024egoplan} for human-level planning. We evaluate our method against proprietary, open-source, and embodied foundation models of comparable scale. Our approach outperforms all baselines, achieving state-of-the-art results. We report the BLEU score for RoboVQA and Accuracy (\%) for EgoPlan-Bench2.}
\label{tab:vlm_vqa}\vspace{-1em}
\end{table*}

\vspace{-1em}
\subsection{Embodied Reasoning and Trajectory Prediction}
\label{sec:exp:general}

We evaluate the embodied reasoning and spatial grounding capabilities of our task manager across four benchmarks: \textbf{RoboVQA}~\cite{sermanet2024robovqa} for long-horizon reasoning, \textbf{EgoPlan-Bench2}~\cite{qiu2024egoplan} for human-level planning, and \textbf{ShareRobot-T}~\cite{robobrain1-0} and \textbf{VABench-V}~\cite{fsd} for 2D trajectory prediction. These capabilities are critical for a task manager to effectively bridge the gap between abstract user instructions and executable robot plans.

Quantitative results are reported in Tab.~\ref{tab:vlm_vqa} and Tab.~\ref{tab:robot_spatial_bench}, where we compare \method against leading proprietary models (e.g., Gemini-3.0-Flash~\cite{gemini3flash}), general-purpose open-source VLMs (e.g., Qwen3-VL~\cite{qwen3vl}), and specialized embodied foundation models of comparable scale(e.g., ThinkAct~\cite{thinkact}). Our method consistently outperforms all baselines, demonstrating superior accuracy in both high-level semantic planning and low-level spatial trajectory generation. 

\begin{figure}[t]
\centering
\includegraphics[width=1\linewidth]{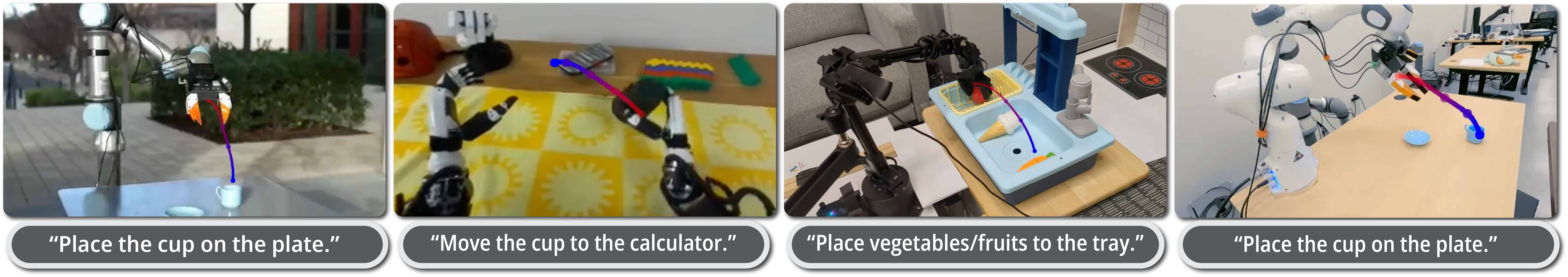}
\vspace{-1em}
\caption{\footnotesize \textbf{Qualitative results of trajectory prediction on out-of-distribution embodiments.} Our high-level task manager generalizes across diverse robotic setups and environments, including different manipulators, objects, and scenes. By decoupling high-level task planning from low-level control, the manager can seamlessly plug into downstream executor policies trained on different embodiments and datasets.}
\label{fig:vlm-cross}\vspace{-1em}
\end{figure}

A key strength of our high-level task manager is its ability to generalize across robotic setups and environments, including varying manipulators, object categories, and scene. The manager acts as a modular component that can seamlessly interface with different downstream executor policies. This architectural decoupling allows \method to remain effective even when deployed on embodiments or datasets not seen during the training of the task manager, highlighting its potential as a foundation model for embodied planning. To further highlight these advantages, we provide qualitative samples in Fig.~\ref{fig:vlm-cross}, showing results of trajectory prediction on out-of-distribution embodiments.

\begin{table*}[t]
\small
\centering
\setlength{\tabcolsep}{0.1pt}
\scalebox{0.97}{ 
{\fontsize{8pt}{9pt}\selectfont
\begin{tabular}{l *{6}{>{\centering\arraybackslash}p{45.5pt}}} 
\toprule
 & \multicolumn{3}{c}{\textbf{ShareRobot-T}} & \multicolumn{3}{c}{\textbf{VABench-V}} \\
\cmidrule(lr){2-4} \cmidrule(lr){5-7}
Methods & DFD $\downarrow$ & HD $\downarrow$ & RMSE $\downarrow$ & DFD $\downarrow$ & HD $\downarrow$ & RMSE $\downarrow$ \\
\midrule
Qwen3-VL-4B~\cite{qwen3vl} & 0.3808 & 0.3294 & 0.2204 & 0.2792 & 0.2528 & 0.2037 \\
MolmoAct-7B~\cite{molmoact} & 0.7764 & 0.7764 & 0.6771 & 0.8136 & 0.8136 & 0.6877 \\
Hamster-13B~\cite{hamster} & 0.4365 & 0.3919 & 0.3554 & 0.2124 & 0.2045 & 0.1825 \\
Embodied-R1-3B~\cite{embodiedr1} & 0.3426 & 0.3002 & 0.2388 & 0.3028 & 0.2588 & 0.2129 \\
\rowcolor{myyellow!60}
\textbf{\method-4B} & \textbf{0.2309} & \textbf{0.2058} & \textbf{0.1559} & \textbf{0.2123} & \textbf{0.1821} & \textbf{0.1469} \\
\bottomrule
\end{tabular}}}
\vspace{0.1em}
\caption{\footnotesize Performance comparison on trajectory prediction benchmarks. We evaluate on ShareRobot-T~\cite{robobrain1-0} and VABench-V~\cite{fsd} across three key metrics: Discrete Fréchet Distance (DFD), Hausdorff Distance (HD), and Root Mean Square Error (RMSE).}
\label{tab:robot_spatial_bench}\vspace{-1em}
\end{table*}

\vspace{-1em}
\subsection{Embodied Agent}
\label{sec:exp:agent}

To evaluate the high-level planning ability of our task manager, we conduct experiments on EmbodiedBench~\cite{embodiedbench}. This benchmark is uniquely designed to evaluate embodied agents' proficiency in instruction understanding, commonsense reasoning, and long-horizon planning. 

In this setup, the model is queried with an egocentric observation image and must output atomic commands (e.g., \texttt{PickUp}, \texttt{Open}) that are directly executable by robot APIs. By bypassing the complexities of continuous low-level motor control while requiring visually-grounded discrete actions, EmbodiedBench serves as the most representative benchmark for VLA capability at the planning level.

We focus our evaluation on two environments:
\begin{itemize}
    \item \textbf{EB-Alfred:} Based on the ALFRED dataset, requiring multi-step household task completion.
    \item \textbf{EB-Habitat:} Utilizing the Habitat simulator for semantic navigation and interaction in 3D indoor scenes.
\end{itemize}

Tab.~\ref{tab:embodied_bench} reports the evaluation results. Our method achieves leading performance across both EB-Alfred and EB-Habitat compared with existing multimodal models of similar or even larger scale. These results demonstrate that \method can effectively interpret long-horizon instructions and produce accurate sub-task plans, highlighting the strength of our hierarchical task reasoning design.

\begin{table*}[t]
    \small
    \centering
    \setlength{\tabcolsep}{3.2pt}
    \scalebox{0.97}{
    {\fontsize{8pt}{9pt}\selectfont
    \begin{tabular}{l *{7}{>{\centering\arraybackslash}p{33pt}}} 
    \toprule
    & \multicolumn{7}{c}{\textbf{EmbodiedBench} $\uparrow$} \\
    \cmidrule(lr){2-8}
    Methods & Base  & Common  & Complex  & Long  & Spatial  & Visual  & Average \\
    \midrule
    \rowcolor{gray!10}
    \multicolumn{8}{l}{\raisebox{-0.2ex}{\includegraphics[height=1em]{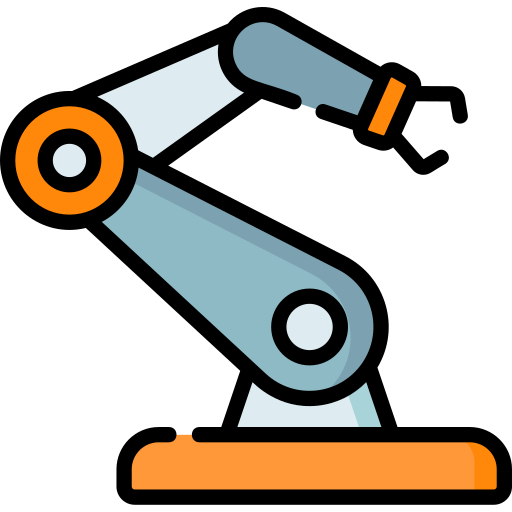}} \bf\textit{EB-Alfred}}\\
    GPT-4o mini~\cite{gpt4omini} & 0.34 & 0.28 & 0.36 & 0.00 & 0.22 & 0.24 & 0.24 \\
    Llama3.2-11B~\cite{llama3-2} & 0.24 & 0.08 & 0.16 & 0.06 & 0.06 & 0.22 & 0.14 \\ 
    InternVL3-8B~\cite{zhu2025internvl3} & 0.20 & 0.14 & 0.14 & 0.02 & 0.00 & 0.12 & 0.10 \\
    Qwen3-VL-4B~\cite{qwen3vl} & 0.18 & 0.14 & 0.26 & 0.26 & 0.14 & 0.14 & 0.19 \\
    \rowcolor{myyellow!60}
    \bf \method-4B & \bf 0.48 & \bf 0.36 & \bf 0.54 & \bf 0.31 & \bf 0.30 & \bf 0.28 & \bf 0.38 \\
    \midrule
    \rowcolor{gray!10}
    \multicolumn{8}{l}{\raisebox{-0.2ex}{\includegraphics[height=1em]{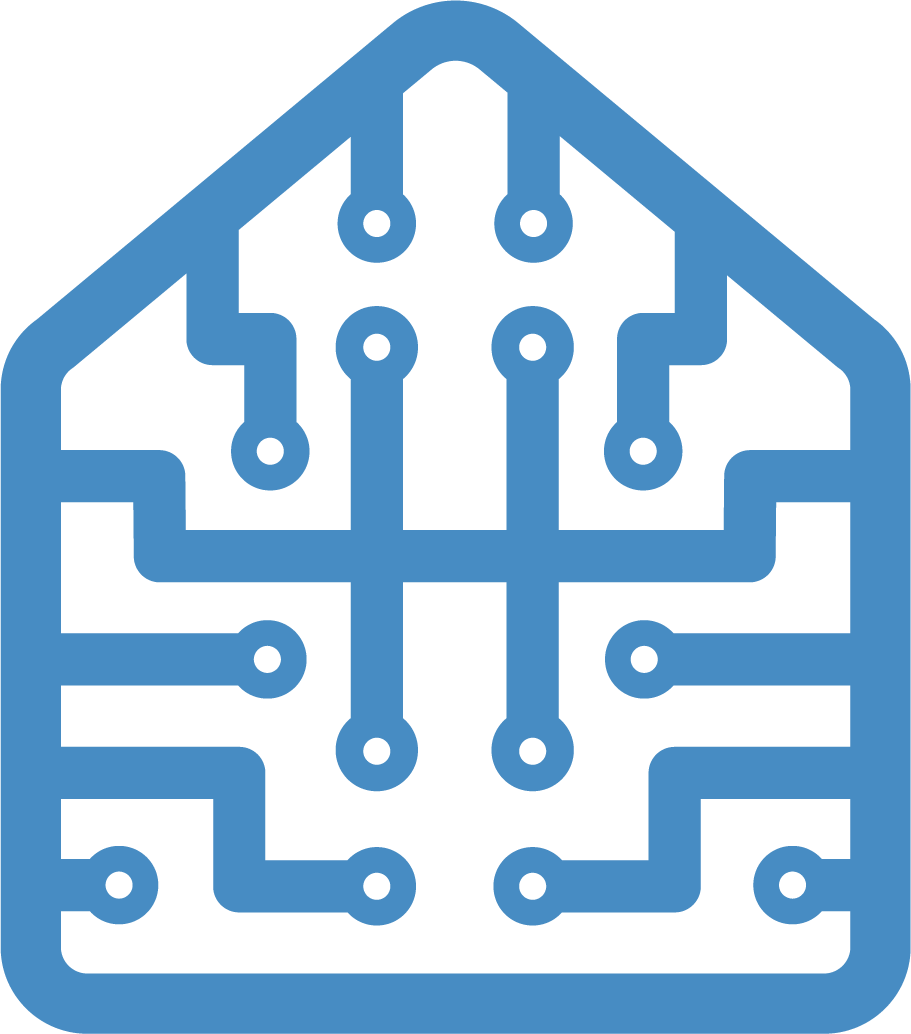}} \bf\textit{EB-Habitat}}\\ 
    GPT-4o mini~\cite{gpt4omini} & 0.74 & 0.22 & 0.32 & 0.14 & 0.32 & 0.22 & 0.33 \\
    Llama3.2-11B~\cite{llama3-2} & 0.70 & 0.16 & 0.28 & 0.06 & 0.20 & 0.10 & 0.25 \\ 
    InternVL3-8B~\cite{zhu2025internvl3} & 0.60 & 0.14 & 0.24 & 0.10 & 0.20 & 0.18 & 0.24 \\
    Qwen3-VL-4B~\cite{qwen3vl} & 0.68 & 0.16 & 0.38 & 0.10 & 0.26 & 0.20 & 0.30 \\
    \rowcolor{myyellow!60}
    \bf \method-4B & \bf 0.78 & \bf 0.32 & \bf 0.46 & \bf 0.20 & \bf 0.28 & \bf 0.26 & \bf 0.38 \\
    \bottomrule
    \end{tabular}}}
    \vspace{0.1cm}
    \caption{\footnotesize Performance comparison on EmbodiedBench~\cite{embodiedbench}. \method substantially outperforms baselines, especially on the Complex configuration.}
    \label{tab:embodied_bench}\vspace{-1em}
\end{table*}

\subsection{VLA in Simulation}
\vspace{-0.5em}
\label{sec:exp:vla-sim}

We evaluate our method across two simulation benchmarks: LIBERO~\cite{liu2023libero} and VLABench~\cite{zhang2025vlabench}, which provide a comprehensive assessment of both foundational manipulation and complex reasoning.

VLABench is specifically designed to challenge the reasoning and generalization capabilities of VLA models, with a primary focus on long-horizon tasks. The benchmark features scenarios where instructions are often implicitly specified and visual observations deviate significantly from the training distribution. This setup necessitates that agents rely on semantic reasoning and robust visual grounding rather than simple memorization of action patterns. In contrast, while LIBERO focuses on rather shorter horizons, it serves as a widely adopted standard for evaluating the atomic capabilities and base motor skills of VLA models. By evaluating on both, we demonstrate our framework's versatility across different task complexities.

We report quantitative results in Tab.~\ref{tab:libero} and Tab.~\ref{tab:vlabench}. Our method consistently outperforms prior approaches across all task suites. For VLABench, we further analyze performance using the Intention Score (IS) and Progress Score (PS), as shown in Fig.~\ref{fig:real-bar}. The Progress Score is particularly meaningful for evaluating long-horizon success, as it captures the agent's ability to complete intermediate steps. Our superior performance on these metrics demonstrates the effectiveness of our hierarchical framework in successfully bridging high-level reasoning with precise low-level control for complex manipulation.

\begin{figure}[t]
\centering
\includegraphics[width=1\textwidth]{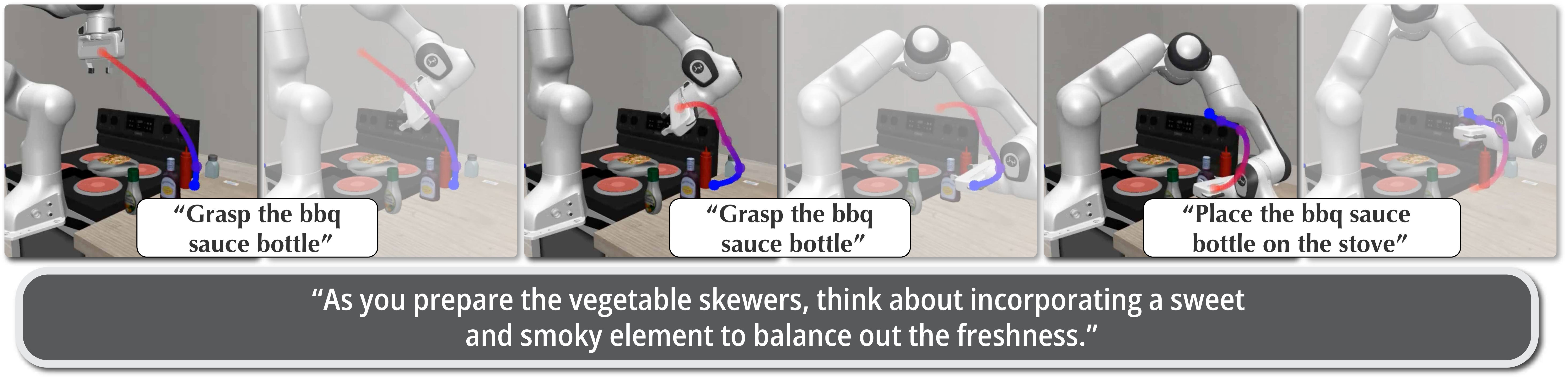}
\vspace{-2em}
\caption{\footnotesize Results on the semantic instruction track of VLABench. The task involves high-level abstract instructions that require reasoning to infer the appropriate manipulation actions. We visualize both the predicted subtasks and the corresponding trajectory generated by our method.}
\label{fig:vlabench-qual}\vspace{-1em}
\end{figure}

To further investigate the reasoning process, we provide qualitative results on VLABench in Fig.~\ref{fig:vlabench-qual}. We visualize the agent's ability to decompose complex, implicit instructions—such as ``\textit{As you prepare the vegetable skewers...}''—into a coherent sequence of executable atomic commands. These visualizations highlight how our method maintains consistent visual grounding throughout long-horizon executions, successfully navigating significant visual shifts and distractor objects that often cause baseline models to fail.

\begin{table*}[t]
    \small
    \centering
    \setlength{\tabcolsep}{4.5pt}
    \scalebox{0.97}{
    {\fontsize{8pt}{9pt}\selectfont
    \begin{tabular}{lcccccc}
    \toprule
    & \multicolumn{6}{c}{\textbf{VLABench}} \\
    \cmidrule(lr){2-7}
    Methods &  
    \begin{tabular}[c]{@{}c@{}}In \\ Distribution $\uparrow$\end{tabular} &  
    \begin{tabular}[c]{@{}c@{}}Cross \\ Category $\uparrow$\end{tabular} &  
    \begin{tabular}[c]{@{}c@{}}Common \\ Sense $\uparrow$\end{tabular} &  
    \begin{tabular}[c]{@{}c@{}}Semantic \\ Instr. $\uparrow$\end{tabular} &  
    \begin{tabular}[c]{@{}c@{}}Unseen \\ Texture $\uparrow$\end{tabular} &  
    \begin{tabular}[c]{@{}c@{}}\textbf{Average} $\uparrow$ \\\end{tabular} \\
    \midrule
    $\pi_0$-fast~\cite{pi0fast} & 0.29 & 0.18 & 0.21 & 0.20 & 0.24 & 0.22 \\
    $\pi_{0.5}$~\cite{pi0-5} & 0.37 & 0.22 & 0.21 & 0.17 & 0.25 & 0.24 \\
    \rowcolor{myyellow!60}
    \bf \method & \bf 0.54 & \bf 0.23 & \bf 0.36 & \bf 0.42 & \bf 0.39 & \bf 0.39 \\
    \bottomrule
    \end{tabular}}}
    \vspace{0.1cm}
    \caption{\footnotesize Performance comparisons on VLABench~\cite{zhang2025vlabench}. \method consistently outperform our base VLA $\pi_{0.5}$.}
    \label{tab:vlabench}\vspace{-1em}
\end{table*}

\begin{table*}[t]
    \small
    \centering
    \setlength{\tabcolsep}{8pt} 
    \scalebox{0.97}{
    {\fontsize{8pt}{9pt}\selectfont
    \begin{tabular}{lccccc}
    \toprule
    & \multicolumn{5}{c}{\textbf{LIBERO}} \\
    \cmidrule(lr){2-6}
    Methods &  
    \begin{tabular}[c]{@{}c@{}}Spatial $\uparrow$\end{tabular} &  
    \begin{tabular}[c]{@{}c@{}}Object $\uparrow$\end{tabular} &  
    \begin{tabular}[c]{@{}c@{}}Goal $\uparrow$\end{tabular} &  
    \begin{tabular}[c]{@{}c@{}}Long $\uparrow$\end{tabular} &  
    \begin{tabular}[c]{@{}c@{}}\textbf{Average} $\uparrow$\end{tabular} \\
    \midrule
    Octo-Base~\cite{octo} & 78.9 & 85.7 & 84.6 & 51.1 & 75.1 \\
    OpenVLA~\cite{openvla} & 84.7 & 88.4 & 79.2 & 53.7 & 76.5 \\
    DiT-Policy~\cite{dasari2025ingredients} & 82.6 & 84.7 & 82.1 & 57.6 & 76.8 \\
    TraceVLA~\cite{zheng2025tracevla} & 84.6 & 85.2 & 75.1 & 54.1 & 74.8 \\
    CoT-VLA~\cite{cotvla} & 87.5 & 91.6 & 87.6 & 69.0 & 83.9 \\
    SpatialVLA~\cite{spatialvla} & 88.2 & 89.9 & 78.6 & 55.5 & 78.1 \\
    Nora-AC~\cite{nora} & 85.6 & 89.4 & 80.0 & 63.0 & 79.5 \\
    WorldVLA~\cite{worldvla} & 87.6 & 96.2 & 83.4 & 60.0 & 79.1 \\
    ThinkAct~\cite{thinkact} & 88.3 & 91.4 & 87.1 & 70.9 & 84.4 \\
    $\pi_0$-fast~\cite{pi0fast} & 96.4 & 96.8 & 88.6 & 60.2 & 85.5 \\
    GR00T-N1.5~\cite{gr00t-n1_5} & 92.0 & 92.0 & 86.0 & 76.0 & 86.5 \\
    MolmoAct~\cite{molmoact} & 87.0 & 95.4 & 87.6 & 77.2 & 86.6 \\
    StarVLA~\cite{community2026starvla} & 97.8 & 98.6 & 96.2 & 93.8 & 96.6 \\
    \rowcolor{myyellow!60}
    \bf \method & \bf 98.0 & \bf 98.6 & \bf 98.0 & \bf 95.2 & \bf 97.5\\
    \bottomrule
    \end{tabular}}}
    \vspace{0.1cm}
    \caption{Performance on LIBERO Benchmark. Our method achieves the highest average score across all four tracks.}
    \vspace{-1em}
    \label{tab:libero}
\end{table*}

\vspace{-1em}

\subsection{VLA in Real-World}
\vspace{-0.5em}
\label{sec:exp:vla-real}

\begin{figure}[t]
\centering
\includegraphics[width=1\textwidth]{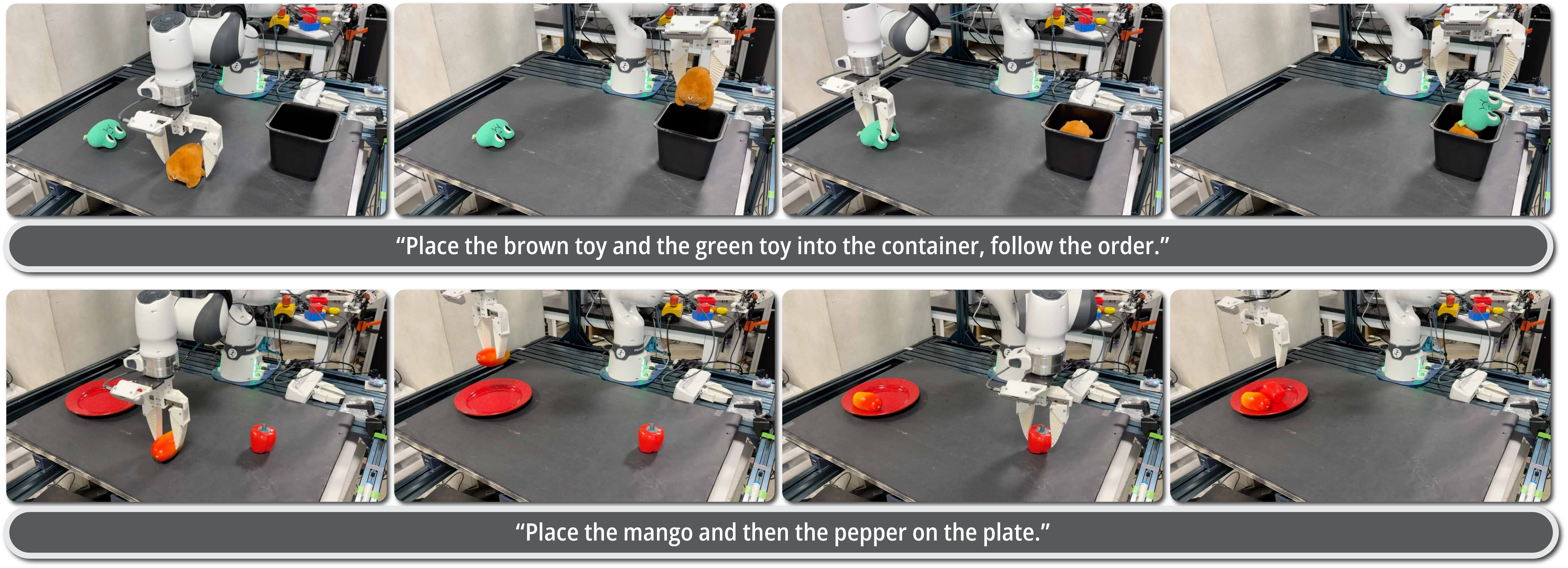}
\vspace{-2em}
\caption{\footnotesize Robot execution results on multi-step tasks. We evaluate performance under both in-distribution and out-of-distribution settings, including scenarios with unseen objects and novel subtask combinations.}
\label{fig:real-qual}\vspace{-1em}
\end{figure}

To evaluate the effectiveness of \method in physical environments, we conduct experiments on a real-world robot setup. Our hardware configuration consists of a Franka arm equipped with two Intel RealSense cameras: one providing a static top-view of the workspace and another mounted on the gripper for wrist-view observations. We collected 100 demonstration trajectories via teleoperation~\cite{yan2025ace}, which were subsequently processed through our automated data pipeline (Sec.~\ref{sec:method}) to fine-tune both the high-level task manager and the low-level action executor.
The evaluation is categorized into single-step and multi-step tasks. Single-step tasks focus on the precision and grounding of a single atomic command, while multi-step tasks require the task manager to maintain temporal consistency and correct sequencing over long horizons. In Fig.~\ref{fig:real-qual}, we provide qualitative visualizations of two multi-step samples, demonstrating how our method successfully executes a chain of sub-goals to complete long-horizon manipulation sequences.

We specifically focus our evaluation on Out-of-Distribution (OOD) scenarios to test the robustness of the decoupled architecture. We define OOD conditions based on the task complexity:
\begin{itemize}
    \item Single-step OOD: Evaluated using novel object categories that were not present in the fine-tuning demonstrations, testing the model's zero-shot semantic grounding.
    \item Multi-step OOD: Evaluated through novel spatial arrangements and unseen language combinations. This requires the task manager to generalize its reasoning to new environmental layouts and instruction structures while maintaining the correct execution logic.
\end{itemize}

\begin{figure}[t]
\centering
\includegraphics[width=0.81\textwidth]{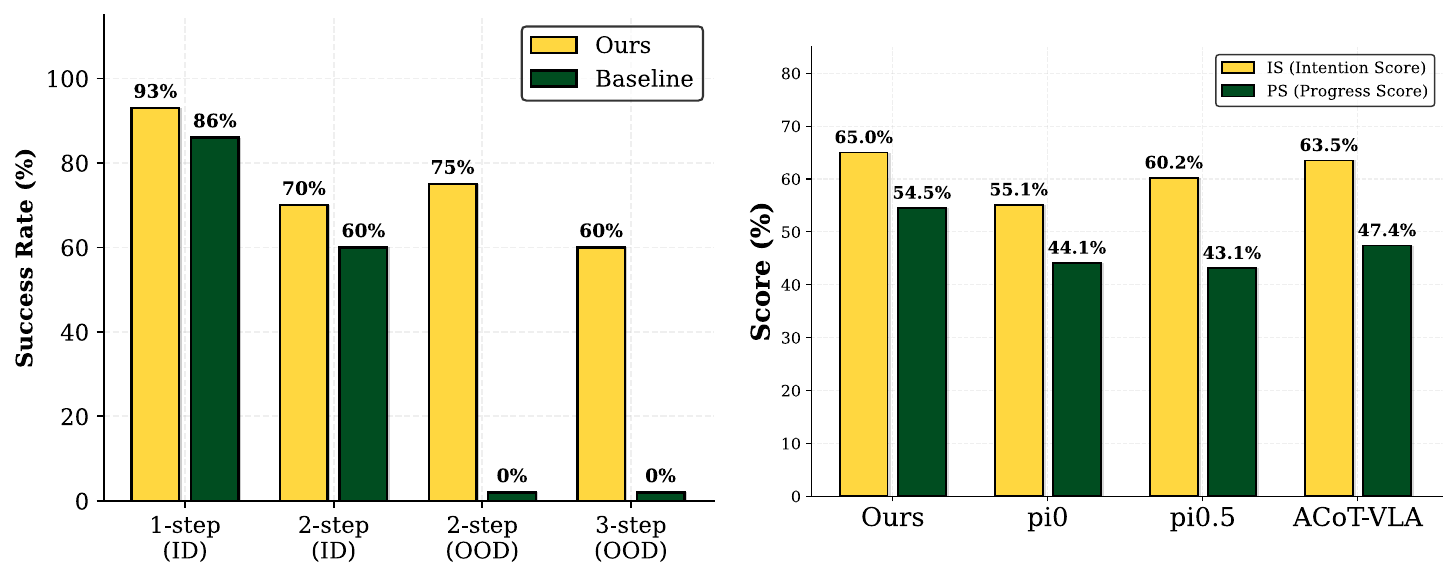}
\caption{\footnotesize (Left) \textbf{Real robot results}. \method significantly outperforms $\pi_{0.5}$ (our base VLA finetuned with the same data) in out-of-distribution settings involving novel instructions and objects, demonstrating superior generalization. (Right) \textbf{IS and PS comparison on VLABench}. \method consistently exceeds baselines in metrics requiring robust vision-language reasoning, moving beyond simple success rates.}
\label{fig:real-bar}\vspace{-1em}
\end{figure}

As shown in Fig.~\ref{fig:real-bar}, our method significantly outperforms our base VLA model, the $\pi_{0.5}$ baseline, which was also fine-tuned on the same 100-sample dataset. The results demonstrate that decoupling high-level semantic planning from low-level execution allows the robot to maintain a higher success rate in OOD settings. In contrast, the baseline model often fails to localize the correct objects or correctly sequence the necessary sub-tasks when faced with novel visual and linguistic arrangements.

\label{sec:exp:abl}

\vspace{-1em}
\section{Related Work}
\label{sec:related-work}

\vspace{-0.5em}
\subsection{Robot Planning with Foundation Models}
Robot planning is traditionally formulated as coupling a discrete task structure with continuous feasibility,
as in task-and-motion planning (TAMP)~\cite{tamp_survey,hpn}.
Recently, foundation models have been used to instantiate the high-level planning layer for open-ended instructions,
including grounded skill selection~\cite{saycan}, embodied multimodal planners~\cite{palme},
and closed-loop feedback with language-state updates~\cite{innermonologue,robix}.
Complementary efforts provide planning-oriented resources, including benchmarks for multimodal planning and long-horizon reasoning~\cite{qiu2024egoplan,sermanet2024robovqa}
and large-scale embodied data/model engines~\cite{robobrain1-0,robobrain2-0}.
A complementary direction treats LLMs as program synthesizers that generate executable control logic
and API calls~\cite{liang2023code,progprompt,reliablecap}.
In parallel, agentic multimodal foundation models study general-purpose interactive decision making for embodied agents~\cite{yang2025magma}.
We also note structured instantiations that couple LLM planning with explicit world models for scalability,
e.g., 3D scene-graph grounded planning in SayPlan~\cite{sayplan}.

\vspace{-1em}
\subsection{Long-Horizon Manipulation and Generalist VLA}
Long-horizon manipulation requires maintaining task state over many steps, handling subtask dependencies,
and recovering from compounding execution errors.
Generalist VLA foundations provide strong short-horizon control and broad transfer~\cite{rt1,rt2,openx,openvla,pi0,pi0-5,pi0fast,gr00t-n1,lbm},
but long-horizon performance is often limited by drift and partial observability.
Therefore, recent work targets long-horizon capability more directly, e.g., by introducing long-horizon training/evaluation
and stronger execution-time correction~\cite{longvla,liu2024bidirectional,cotvla,thinkact,fastthinkact},
or by integrating planning-style intermediate structure into a unified VLA loop~\cite{bagelvla,lingbotvla}.
Another key ingredient is \emph{memory} for progress tracking under partial observability: RoboVQA and MindExplore study
multimodal long-horizon reasoning with explicit mechanisms to integrate observations over time~\cite{sermanet2024robovqa,mindexplore},
and concurrent with our work, MEM equips VLAs with multi-scale memory via short-term visual context and longer-horizon language memory~\cite{mem}. Our work contributes a complementary axis: rather than solely scaling an end-to-end VLA or embedding all reasoning within it,
we separate long-horizon task management from short-horizon control and train the manager to predict \emph{remaining} plans from the current observation.
This receding-horizon ``what remains'' formulation provides a stable progress signal for closed-loop execution, and interfaces naturally with
trace-conditioned policies for local control.

\vspace{-1em}
\subsection{Visual Prompting and Trajectory Representations}
Several lines of work bridge high-level intent to low-level policies by introducing explicit intermediate guidance.
VIMA formulates manipulation as prompt-conditioned behavior via interleaved visual and textual tokens~\cite{vima}.
Planning-oriented systems also use intermediate representations (e.g., code, value maps, or waypoints) to connect
semantic decisions to short-horizon execution~\cite{liang2023code,voxposer,thinkact}.
More directly, TraceVLA encodes trajectories as a visual trace prompt to enhance spatial-temporal awareness for generalist robotic policies~\cite{zheng2025tracevla}.
Our approach follows this theme by using a lightweight \emph{visual trace} as a direct conditioning signal,
reducing long-horizon manipulation into repeated short-horizon, locally-conditioned execution.

\vspace{-1em}
\section{Conclusion}
\vspace{-0.5em}
\label{sec:conclusion}

We present \method, a hierarchical framework for long-horizon manipulation that decouples high-level symbolic task planning from low-level sensorimotor execution. By predicting a joint distribution of sub-task sequences and canonical visual traces, our approach reformulates complex long-horizon reasoning into a series of manageable, short-horizon control problems. Extensive experiments across embodied QA, simulation, and real-world benchmarks demonstrate that \method significantly improves success rates, generalization, and task-planning accuracy compared to monolithic VLA baselines. Our architecture provides a scalable pathway for integrating future advancements in both large-scale VLMs and high-frequency robot control policies.

\clearpage
\appendix
\clearpage
\appendix

\definecolor{cvprblue}{rgb}{0.21,0.49,0.74}

\setcounter{section}{0}
\renewcommand{\thesection}{\Alph{section}}

\titlecontents{section}
  [0em]
  {\color{cvprblue}}
  {\contentslabel{2.3em}}
  {}
  {\titlerule*[0.5pc]{.}\contentspage}

\titlecontents{subsection}
  [2.3em]
  {\color{cvprblue}}
  {\contentslabel{2.8em}}
  {}
  {\titlerule*[0.5pc]{.}\contentspage}

\startcontents[supp]

\begin{center}
    {\Large\bfseries Supplementary Material\par}
\end{center}
\vspace{0.5em}

{
\setcounter{tocdepth}{2}
\printcontents[supp]{l}{1}{}
}

\section{More Qualitative Results on VLABench}
\label{sec:vlabench}

We provide additional qualitative results of our method on VLABench~\cite{zhang2025vlabench}.
The VLABench contains five evaluation tracks, each consisting of 50 tasks with different scenes or instructions.

\begin{itemize}
    \item \textbf{In-distribution}. Evaluates the policy’s ability to learn tasks from in-domain episodes with a small and diverse dataset.
    \item \textbf{Cross-category}. Evaluates generalization across different object categories and instances.
    \item \textbf{Common-sense}. Evaluates the ability to apply common-sense reasoning to identify the target object.
    \item \textbf{Semantic-instruction}. Evaluates the understanding of instructions with complex semantic or contextual information.
    \item \textbf{Unseen-texture}. Evaluates robustness under changes in background and table textures.
\end{itemize}

\begin{figure}[p]
\centering
\includegraphics[width=\textwidth]{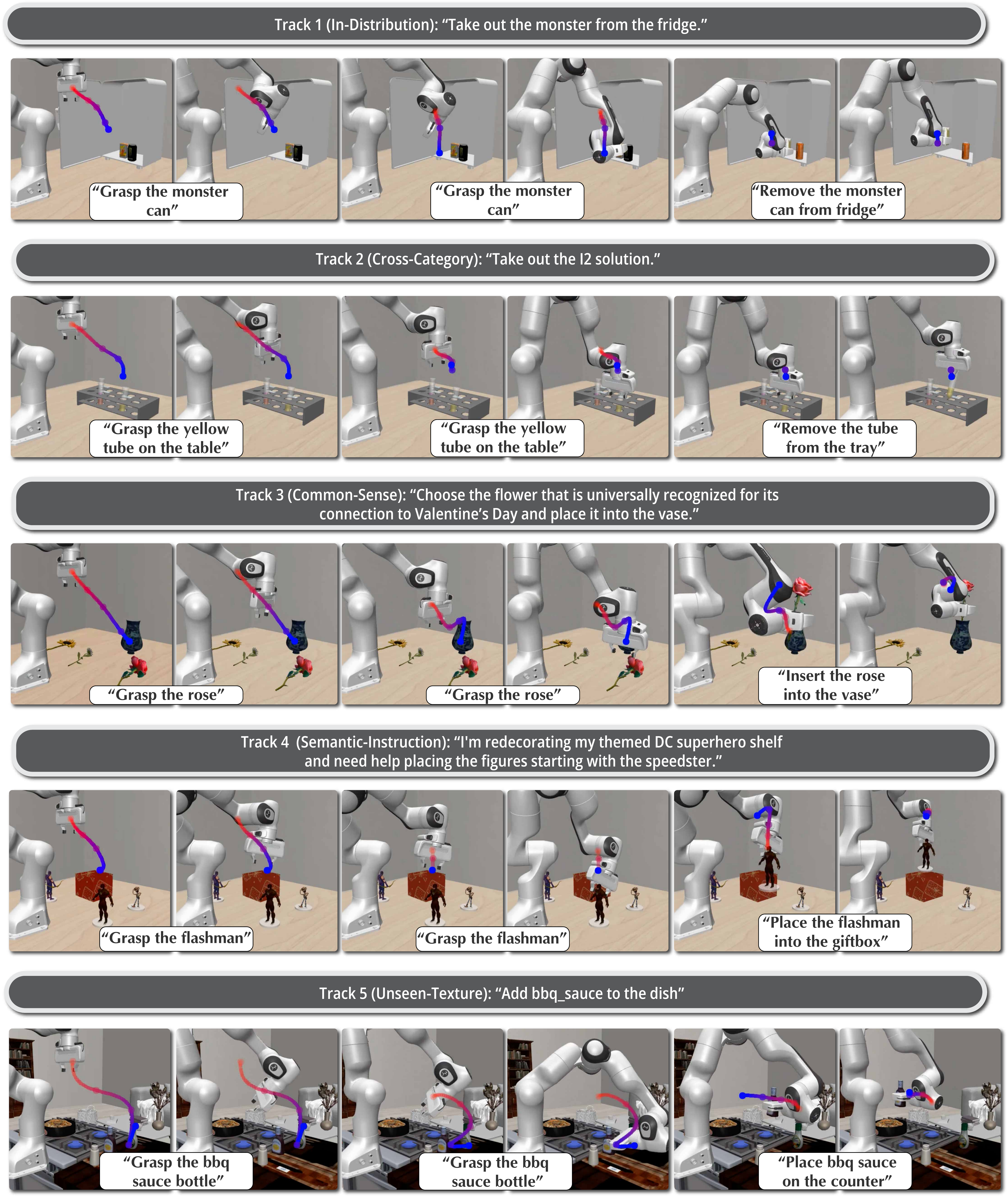}
\caption{\footnotesize Results on different tracks of VLABench. We show the predicted sub-tasks and their corresponding trajectory.}
\label{fig:supp_vlabench}
\end{figure}

Results from Fig.~\ref{fig:supp_vlabench} demonstrate that our method generalizes well to both in-distribution and out-of-distribution scenes and tasks. For the common-sense and semantic-instruction tracks, which involve open-ended instructions, our method shows strong reasoning and generalization ability. For the unseen-texture track with varying textures, our method remains robust and continues to generate reasonable sub-tasks and actions.

\section{More Qualitative Results on Real Evaluation}
\label{sec:real-eval}

We provide additional qualitative results from real robot experiments. The system takes a top-down camera view, a wrist-mounted camera view, and a high-level instruction as input. The task manager first predicts the next sub-task together with a corresponding visual trace. The VLA executor then generates low-level robot actions conditioned on these signals to perform the manipulation.

The results demonstrate that our framework can reliably decompose high-level instructions into executable sub-tasks and produce spatial traces that guide the robot toward the correct interaction targets. The predicted traces provide an intuitive intermediate representation that improves the stability of execution and helps the robot recover from minor deviations during long-horizon manipulation. Across different scenes and object configurations, the system maintains consistent task progress and successfully completes multi-step manipulation sequences.

\begin{figure}[ht]
\centering
\includegraphics[width=\textwidth]{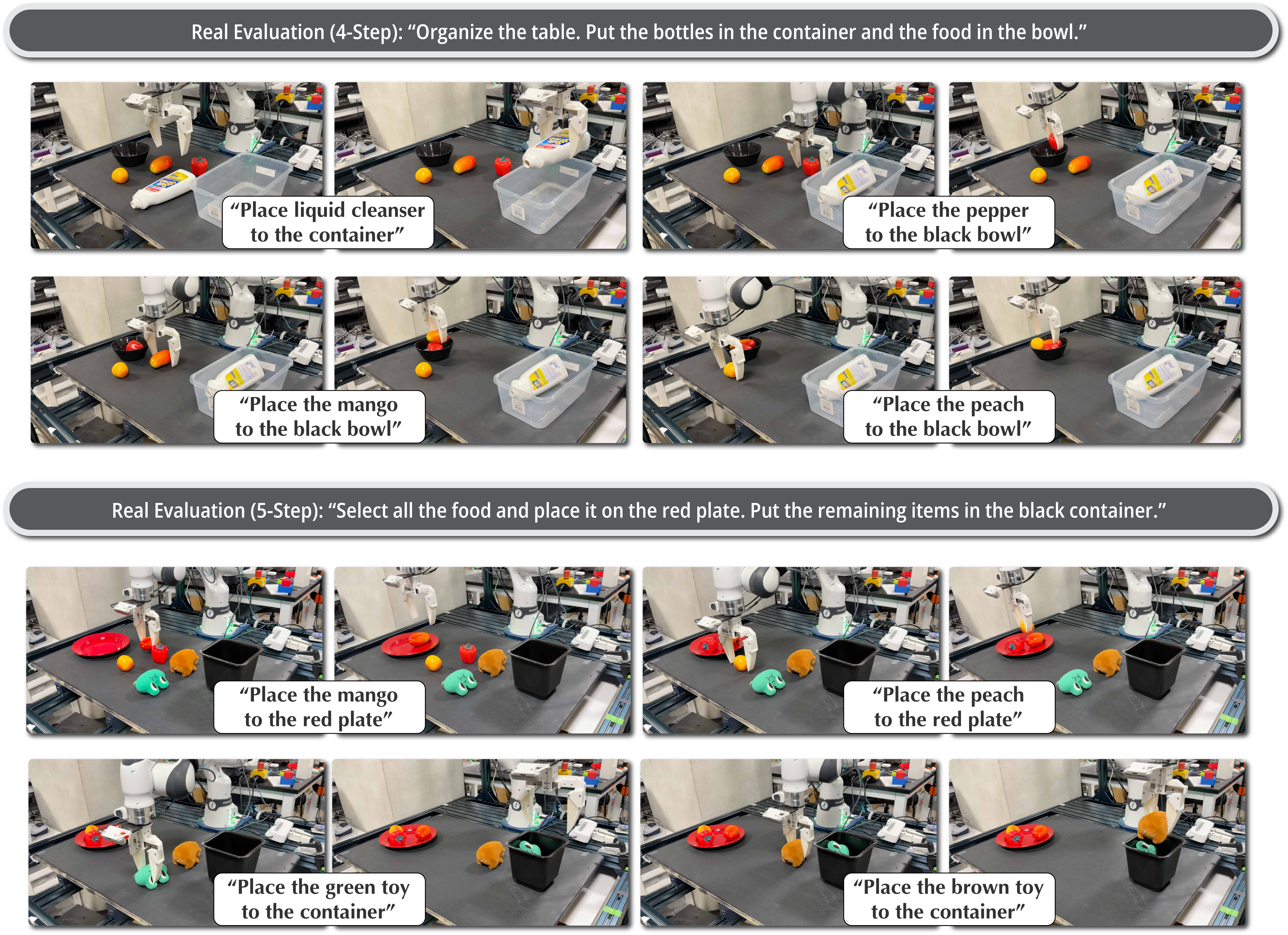}
\caption{\footnotesize Additional qualitative results from real robot experiments. We evaluate our model on different long-horizon manipulation tasks.}
\label{fig:supp_real}
\end{figure}

\section{More Ablation Study}
\label{sec:ablation}
\subsection{Quality of Sub-Task and Trace Curation}

We study the effectiveness of the proposed sub-task and trace construction in our data pipeline for generating training data from real robot demonstrations. We compare the performance before and after incorporating our curated data on trajectory prediction benchmarks on ShareRobot-T~\cite{robobrain1-0} and VABench-V~\cite{fsd} across three key metrics: Discrete Fréchet Distance (DFD), Hausdorff Distance (HD), and Root Mean Square Error (RMSE). 

The results in Tab.~\ref{tab:supp_traj_ablat} demonstrate that our method greatly improved the model's trajectory prediction ability on both benchmarks, validating the effectiveness of our curated sub-task and trajectory data.

\begin{table*}[h]
\small
\centering
\setlength{\tabcolsep}{0.1pt}
\scalebox{0.97}{ 
{\fontsize{8pt}{9pt}\selectfont
\begin{tabular}{l *{6}{>{\centering\arraybackslash}p{45.5pt}}} 
\toprule
 & \multicolumn{3}{c}{\textbf{ShareRobot-T}} & \multicolumn{3}{c}{\textbf{VABench-V}} \\
\cmidrule(lr){2-4} \cmidrule(lr){5-7}
Methods & DFD $\downarrow$ & HD $\downarrow$ & RMSE $\downarrow$ & DFD $\downarrow$ & HD $\downarrow$ & RMSE $\downarrow$ \\
\midrule
W./o. subtask traj. & 0.2437 & 0.2149 & 0.1628 & 0.2500 & 0.1984 & 0.1686 \\
Ours & 0.2309 & 0.2058 & 0.1559 & 0.2123 & 0.1821 & 0.1469 \\
\bottomrule
\end{tabular}}}
\vspace{0.2em}
\caption{\footnotesize Ablation on trajectory prediction benchmarks before and after incorporating our curated sub-task and trace data. Our data curation pipeline consistently improves trajectory prediction performance on both ShareRobot-T and VABench-V across all metrics (DFD, HD, RMSE).}
\label{tab:supp_traj_ablat}
\end{table*}

\subsection{Integration of Different VLA Executors}

Since our framework adapts a modular design that decouples the high-level task manager from the low-level VLA executor. The task manager predicts progress-aware sub-tasks together with a spatial trace, which serves as a general interface that can be followed by different VLA policies. This design allows the same task manager to be reused with various executor architectures without modifying the planning component.

To validate this property, we integrate our task manager with an alternative VLA model, StarVLA\cite{community2026starvla}, replacing the default $\pi_{0.5}$ executor used in our main experiments. The predicted sub-tasks and visual traces from the manager are provided to the StarVLA policy as execution guidance, enabling the full pipeline to operate in the same closed-loop manner. We evaluate this configuration on the VLABench benchmark and report the results below in Tab.~\ref{tab:supp_vlabench}. The results show that our task manager can effectively interface with different VLA executors, demonstrating the flexibility and extensibility of our framework.

\begin{table*}[ht]
    \small
    \centering
    \setlength{\tabcolsep}{4.0pt}
    \scalebox{0.97}{
    {\fontsize{8pt}{9pt}\selectfont
    \begin{tabular}{ccccccc}
    \toprule
    & \multicolumn{6}{c}{\textbf{VLABench}} \\
    \cmidrule(lr){2-7}
    Methods &  
    \begin{tabular}[c]{@{}c@{}}In \\ Distribution $\uparrow$\end{tabular} &  
    \begin{tabular}[c]{@{}c@{}}Cross \\ Category $\uparrow$\end{tabular} &  
    \begin{tabular}[c]{@{}c@{}}Common \\ Sense $\uparrow$\end{tabular} &  
    \begin{tabular}[c]{@{}c@{}}Semantic \\ Instr. $\uparrow$\end{tabular} &  
    \begin{tabular}[c]{@{}c@{}}Unseen \\ Texture $\uparrow$\end{tabular} &  
    \begin{tabular}[c]{@{}c@{}}\textbf{Average} $\uparrow$ \\\end{tabular} \\
    \midrule
    StarVLA & 0.30 & 0.08 & 0.19 & 0.15 & 0.18 & 0.18 \\
    StarVLA + ours & 0.42 & 0.12 & 0.18 & 0.21 & 0.25 & 0.24 \\
    \bottomrule
    \end{tabular}}}
    \vspace{0.1cm}
    \caption{\footnotesize Performance comparison on VLABench when integrating our task manager with different VLA executors. By providing sub-task and trace guidance, our framework consistently improves the performance of the StarVLA policy across all evaluation tracks.}
    \label{tab:supp_vlabench}\vspace{-1em}
\end{table*}

\subsection{Inference Efficiency}
\label{sec:supp:efficiency}

Our framework maintains efficient inference through a hierarchical execution scheme. In practice, the task manager (planner) is invoked once every 100 steps of the low-level VLA executor, while the executor continuously produces robot actions conditioned on the predicted sub-task and visual trace. Since the planner runs at a much lower frequency than the executor, the additional computational overhead is minimal, keeping the overall inference time comparable to using the executor alone.

We compare the inference latency in the same simulation task before and after integrating the task manager, as shown in Tab.~\ref{tab:supp_latency}.

\begin{table}[h]
\centering
\small
\begin{tabular}{lc}
\toprule
Configuration & Latency (1 episode) ($\downarrow$) \\
\midrule
With task manager & $\sim$86s \\
Without task manager & $\sim$72s \\
\bottomrule
\end{tabular}
\vspace{0.2em}
\caption{Inference latency comparison for a single episode rollout in simulation.}
\label{tab:supp_latency}
\end{table}

In real robot deployment, on a machine equipped with an NVIDIA A6000 GPU, the planner runs at approximately 2\,Hz, while the low-level executor operates at about 10\,Hz. This scheduling allows the system to perform high-level reasoning intermittently while maintaining fast low-level control during long-horizon manipulation.

\section{Details for Sub-Task and Trace Extraction}
\label{sec:supp:trace_prompt}

\begin{table}[p]
\centering
\small
\begin{tabular}{|p{0.15\linewidth}|p{0.75\linewidth}|}
\hline
Data Type & Prompt Template \\ \hline
Robot Trace Detection & You are an expert in robotic vision. Locate the Franka robot's end-effector in this top-down image. \\
& \#\#\# Target Definition \\
& Detect the entire end-effector assembly only. This includes the wrist (hand) and the grippers/fingers. Do NOT include the forearm (Link 7) or any part of the articulated arm leading up to the wrist. \\
& \#\#\# Strict Output Format \\
& Return a valid JSON array of exactly one object, or an empty array [] if the end-effector is not visible. \\
& Format: [\{"bbox\_2d": [xmin, ymin, xmax, ymax], "label": "robot\_end\_effector"\}] \\
& \#\#\# Rules \\
& - Coordinates must be integers in pixel units (Origin: top-left). \\
& - The box must be the tightest possible fit around the grippers and wrist housing. \\
& - Output ONLY the JSON. No markdown code blocks, no preamble, no commentary. \\ \hline
Sub-task Decomposition & You are an expert in robot action analysis. Your task is to: \\
& 1. Decompose the robot's behavior into atomic sub-tasks that directly fulfill the instruction: '\{language\}'. \\
& \#\#\# Step 1: Identify Atomic Sub-Tasks \\
& - List only atomic sub-tasks that involve a physical interaction changing an object’s contact or state (e.g., grasp, place, insert, push, cut). \\
& - Exclude: motion-only actions (e.g., move, reach, approach), preparatory phrases (e.g., prepare to...), or non-interactive steps (e.g., release, hover). \\
& - Each sub-task must correspond to one clear change in object state or contact. \\
& - For each sub-task, specify the frame number at which the action is fully completed. \\
& \#\#\# Step 2: Output Format \\
& Return a valid JSON object with following keys: \\
& - 'subtasks': a dictionary mapping each sub-task (string) and its completion frame (as a stringified integer). \\
& \#\#\# Example Output \\
& \{ "subtasks": \{ "10": "Grasp the knife", "25": "Place the knife on the cutting board" \} \} \\ \hline
\end{tabular}
\caption{\footnotesize Prompt templates used for extracting robot end-effector traces and decomposing demonstrations into atomic sub-tasks. A vision-language model is prompted to localize the robot gripper in each frame and to segment manipulation sequences into interaction primitives, producing the trace and sub-task supervision used for training.}
\label{tab:supp_prompt}
\end{table}

To generate the visual trace supervision described in Sec. 2.2, we extract the robot's end-effector
trajectory from demonstration videos using a Vision-Language Model (VLM).

This process converts raw video frames into a sequence of normalized 2D pixel coordinates. For each frame, we query the VLM to localize the Franka robot's end-effector. The prompt defines the target specifically as the gripper and wrist assembly, excluding upstream arm links, to ensure spatial consistency. The model is constrained to a strict JSON output format, returning a bounding box for the detected region. We calculate the center point of this box to determine the robot's location.

The resulting coordinate sequence is normalized to a [0, 1000] range and resampled to create the compact visual trace representation used for training. Frames where the model fails to provide a valid detection are discarded. The prompting template used for this process is detailed in Tab.~\ref{tab:supp_prompt}.
\section{Limitations}
While our method demonstrates strong performance across multiple benchmarks, several aspects remain open for further improvement. 
First, our approach relies on accurate sub-task grounding and trace prediction from the task manager. 
Although the receding-horizon design allows the system to adapt when execution deviates from the plan, improvements in perception and spatial grounding could further enhance the stability of the overall pipeline.

Second, the visual trace representation provides a simple and effective interface between high-level reasoning and low-level control. 
However, it currently captures spatial intent using a 2D trajectory, which may not fully express more complex interaction patterns such as highly precise manipulation or contact-rich behaviors. 
Incorporating richer spatial or temporal guidance signals could further extend the applicability of this representation.

Finally, our experiments primarily focus on tabletop manipulation scenarios with a single robotic arm. 
While the modular design of our framework allows the task manager to interface with different executors, evaluating the approach on a broader range of embodiments, tasks, and dynamic environments would be an interesting direction for future work.

%
%
\clearpage
\bibliographystyle{splncs04}
\bibliography{main}

\end{document}